\documentclass{article}
\usepackage{arxiv}

\usepackage[utf8]{inputenc}
\usepackage[T1]{fontenc}
\usepackage[english]{babel}
\usepackage{url}
\usepackage{booktabs}
\usepackage{amsmath}
\usepackage{amssymb}
\usepackage{amsfonts}
\usepackage{graphicx}
\graphicspath{{figures/}}
\usepackage{array}
\usepackage{nicefrac}
\usepackage{microtype}
\usepackage{hyperref}
\usepackage{placeins}

\title{Lesion Detection in CT with Frozen Self-Distilled Features: SALT, a Spatially Adaptive Label-Guided Temperature}

\author{%
  Mahmut S.~Gokmen\thanks{Corresponding author. TODO(authors): add e-mail addresses and ORCIDs.} \\
  University of Kentucky \\ Lexington, KY, USA
  \And
  Evan W.~Damron \\ University of Kentucky \\ Lexington, KY, USA
  \And
  Mitchell A.~Klusty \\ University of Kentucky \\ Lexington, KY, USA
  \And
  Caroline N.~Leach \\ University of Kentucky \\ Lexington, KY, USA
  \And
  Emily B.~Collier \\ University of Kentucky \\ Lexington, KY, USA
  \And
  V.~K.~Cody Bumgardner \\ University of Louisville \\ Louisville, KY, USA
}

\renewcommand{\shorttitle}{SALT: Spatially Adaptive Label-Guided Temperature}
\hypersetup{
  pdftitle={Lesion Detection in CT with Frozen Self-Distilled Features: SALT},
  pdfauthor={Gokmen, Damron, Klusty, Leach, Collier, Bumgardner},
  pdfkeywords={self-supervised learning, self-distillation, weak supervision, lesion detection, computed tomography}
}

\begin{document}
\maketitle

\begin{abstract} Self-supervised pretraining objectives are spatially uniform: the teacher temperature and the per-patch loss weight are identical everywhere in the image, so a lesion a few patches wide contributes no more to the training signal than the surrounding parenchyma. Prior work biases the \emph{views} toward annotated regions, which changes what the model sees but adds no pressure on the objective. We instead condition the \emph{targets} of self-distillation, a method we call SALT (Spatially Adaptive Label-guided Temperature). Weak, box-derived labels, available \emph{only during pretraining}, define a compact region on the encoder's patch grid, inside which the teacher's softmax temperature is sharpened and the masked-patch loss is up-weighted. The objectives, the masking policy and the centering statistics are otherwise unchanged, and at every downstream use the encoder is a plain feature extractor with no labels and no conditioning. We evaluate by freezing the encoder and training only a lightweight multi-depth CenterNet-style head~\cite{centernet}, detecting lesions in $3$D on four CT cohorts, and we isolate the mechanism against a backbone identical in architecture, pretraining data, schedule and label-guided cropping but with no target conditioning. We report patch-level separability, $3$D detection stratified by cohort and by lesion size, box quality, and a detector-free probe in which a single frozen patch embedding re-identifies a lesion in a follow-up scan without registration, masks or fine-tuning. Because the conditioning is expressed through a spatial indicator rather than through label semantics, the formulation admits any weak spatial annotation; we instantiate and validate it for lesions. \end{abstract}


\section{Introduction}
\label{sec:intro}

Finding lesions in computed tomography underpins both cancer screening and the assessment of treatment response. Low-dose CT screening reduces lung-cancer mortality~\cite{nlst}, and in oncologic follow-up the RECIST~1.1 protocol requires a radiologist to locate, select and measure target lesions at every time point~\cite{recist}. Both tasks depend on finding small, low-contrast structures in volumes of several hundred slices, and both are error-prone: reported miss rates for small pulmonary nodules on chest CT remain close to half of those present, driven largely by lesion conspicuity rather than by image quality~\cite{missednodules}. The regime that matters is therefore the small one, and it dominates the data: $81.6\%$ of the $3983$ lesions in our evaluation are under $6$\,mm in long-axis diameter. Fully supervised $3$D detectors address this task directly but require dense annotations for every anatomy and cohort of interest, which is precisely what is scarce. This motivates asking how much of the burden can be shifted to pretraining, where labels are needed once and never again at inference.

Self-supervised pretraining is now the default route to reusable image encoders, and radiology-specific variants exist at considerable scale~\cite{meddinov3,curia,curia2}. Yet the objectives these models optimize are \emph{spatially uniform}. In self-distillation the teacher's target distribution is sharpened by a single global temperature and every patch enters the loss with the same weight, so a lesion occupying a handful of patches exerts no more influence on the representation than the parenchyma around it. Nothing in the objective asks the encoder to encode it distinctly.

The consequence is visible in the frozen features themselves. A lesion-versus-background direction is linearly decodable from every backbone we evaluate, medical or natural, small or large, at near-ceiling accuracy (AUROC $\ge0.99$). What differs sharply is how \emph{geometrically concentrated} that direction is: for generic encoders, background tissue responds to a lesion prototype almost as strongly as lesion tissue does, and the model with the highest probe accuracy in our comparison also has one of the weakest separation margins (Table~\ref{tab:sep}). Linear decodability is therefore not the same property as a lesion-centered representation, and it is the latter that a lightweight downstream head can exploit.

One established way to bias pretraining toward a structure of interest is to bias the \emph{views}. In earlier work we introduced DINO-LG~\cite{dinolg}, which guides random cropping toward annotated regions so that the target structure is over-represented among the crops, guidance at the level of \emph{what the model sees}. We find this insufficient in the regime studied here. Restricting crops narrows the training distribution but adds no pressure on the objective, and on its own it yields the least concentrated lesion representation of any backbone we evaluate, below both generic CT and natural-image pretraining (Table~\ref{tab:sep}). Where the structure of interest is small and low-contrast, showing the model more of it does not by itself make the model represent it differently.

We therefore move the guidance from the view to the \emph{target}, and call the resulting recipe SALT, for spatially adaptive label-guided temperature. Weak labels, bounding boxes converted to masks, available \emph{only during pretraining}, are pooled onto the encoder's patch grid and reduced to a compact region; inside that region the teacher's softmax temperature is sharpened and the masked-patch loss is up-weighted. The student is thus required to commit to a more decisive code exactly where lesion tissue lies, while the objectives themselves, the masking policy and the centering statistics remain untouched. At every downstream use the encoder is a plain feature extractor: no labels, no conditioning, no fine-tuning. Because the guidance is expressed through a spatial indicator function rather than through label semantics, the formulation is not specific to lesions, any weak spatial annotation defines a valid region, but we instantiate and validate it for lesions only.

We evaluate by freezing the encoder and training only a compact detection head, a protocol that attributes differences in detection to the pretraining objective rather than to downstream capacity. Our contributions are:

\begin{enumerate}\itemsep2pt
\item \textbf{Target-level label guidance.} A conditioning mechanism that modulates the self-distillation \emph{target} through a weak-label-derived spatial indicator: a sharpened teacher temperature and an up-weighted patch loss on a compact region. The conditioning is confined to pretraining and leaves the objectives, the masking and the centering statistics unchanged.
\item \textbf{A controlled account of what it contributes.} Against a backbone identical in architecture, pretraining data, schedule, seed and label-guided cropping, differing \emph{only} in the target conditioning, the mechanism raises the lesion/background separation margin from $0.105$ to $0.449$ and lifts pooled $3$D detection by $+0.11$ CPM, concentrated on sub-$6$\,mm lesions ($+0.14$ CPM) and, across cohorts, carried by two of the four.
\item \textbf{A frozen-encoder $3$D lesion detector.} A lightweight multi-depth CenterNet-style head over frozen patch tokens, evaluated on four CT cohorts ($3983$ lesions) against four external frozen backbones under one shared protocol, with metric definitions, operating points and their selection stated explicitly.
\item \textbf{Representation-level and longitudinal evidence.} A patch-geometry analysis that isolates \emph{why} the mechanism helps, and a detector-free demonstration that the frozen features re-identify a lesion in a follow-up scan without registration, masks or fine-tuning. \end{enumerate}

Section~\ref{sec:related} situates the mechanism among other ways of steering self-supervised objectives with side information. Section~\ref{sec:pretrain} defines the conditioning and Section~\ref{sec:detect} the frozen-encoder detector used to probe it; Section~\ref{sec:exp} reports the data, the protocol and the results, and Section~\ref{sec:discussion} states what they do and do not establish.

\section{Related work}
\label{sec:related}

\subsection{Self-supervised pretraining for radiology}
Medical imaging has followed the general trajectory of self-supervision: contrastive objectives adapted from SimCLR and MoCo~\cite{simclr,moco}, reconstruction-based masked modeling in the manner of MAE~\cite{mae}, task-designed pretext objectives such as Models Genesis~\cite{modelsgenesis}, and, more recently, self-distillation~\cite{dino,ibot,dinov2}. Within radiology this line runs from transfer-learning corpora such as RadImageNet~\cite{radimagenet}, through volumetric pretext pretraining of Swin transformers for $3$D medical analysis~\cite{swinunetr}, to encoders trained at foundation-model scale: MedDINOv3~\cite{meddinov3} adapts a DINOv3~\cite{dinov3} recipe to $\approx\!3.9$M public CT slices for segmentation, while Curia and Curia-2~\cite{curia,curia2} train DINOv2 encoders on $\approx\!200$M CT and MR slices from a private hospital corpus. These models are typically evaluated on classification and segmentation; detection, and localization of small structures in particular, is comparatively under-reported.

Our claim is not that such encoders see too little data, two of our baselines see far more CT than we do, but that their objective is indifferent to \emph{where} the clinically relevant signal lies. The comparison in Table~\ref{tab:sep} is designed to test exactly that: whether more CT pretraining produces a more concentrated lesion representation, or merely a linearly decodable one.

\subsection{Lesion detection in CT}
Universal lesion detection was established by DeepLesion~\cite{deeplesion} and the detector family built on it: 3DCE~\cite{3dce} fuses context from neighboring slices, MULAN~\cite{mulan} adds tagging and segmentation heads to a Mask R-CNN backbone, and LENS~\cite{lens} learns from several partially labeled datasets at once. Task-specific detectors remain strong references: nnDetection~\cite{nndet} self-configures a supervised $3$D pipeline, and the LUNA16 lung-nodule challenge~\cite{luna16} fixed the FROC/CPM evaluation convention that we follow, with the operating-point caveat noted in Sec.~\ref{sec:exp}.

These methods fine-tune an entire network on detection labels. Here the encoder is instead frozen and only a head of a few million parameters is trained, identically for every backbone. The setting is therefore a \emph{representation probe} rather than a bid for state-of-the-art detection, and the comparison of interest is between backbones under a fixed head, not between our absolute numbers and those of a fully fine-tuned detector.

\subsection{Steering self-supervised objectives with side information}
Side information has been injected into self-supervised training at three distinct points, and the distinction matters for placing our contribution.

At the \emph{sampler}, guidance changes which views are seen. CAST~\cite{cast} constrains random crops to overlap an unsupervised saliency map; ContrastiveCrop~\cite{contrastivecrop} places crops using a semantic heatmap; and our own DINO-LG~\cite{dinolg} uses annotations rather than saliency to steer cropping toward a target structure. At the \emph{loss}, guidance changes which pairs or locations are weighted: supervised contrastive learning~\cite{supcon} uses labels to define positives, and saliency-based schemes reweight contrastive terms by how salient the crops are. At the \emph{head}, auxiliary supervised branches are attached alongside the self-supervised objective.

A fourth possibility, modulating the \emph{temperature} of the distillation target, has been explored in supervised teacher-to-student knowledge distillation, but never, to our knowledge, at the granularity we use here. Adaptive-temperature schemes in that literature operate per training step or per example: curriculum temperature anneals the difficulty of the distillation signal over training~\cite{ctkd}, and related methods assign each sample its own temperature from the teacher's confidence or logit scale. Spatial adaptivity does appear in distillation for object detection, but what varies spatially is the masking and the loss weight rather than the target's sharpness; SAMKD, for instance, masks and reweights regions by teacher-student feature discrepancy~\cite{samkd}.

Our mechanism differs from all of these on three axes at once. It is \emph{per token}: the temperature is a function $\tau(p)$ of position on the patch grid, applied to the teacher's target distribution for each masked patch in the iBOT objective, so different tokens of the same image receive targets of different sharpness. It operates inside \emph{self}-distillation, where the teacher is an exponential moving average of the student and the temperature is not merely a smoothing knob but part of the machinery that prevents collapse, which is why we take care to leave the centering statistics unconditioned (Sec.~\ref{sec:pretrain}). And it is driven by an \emph{external weak label} rather than by an intrinsic quantity such as confidence, entropy or a training schedule. We are not aware of prior work combining any two of these, let alone all three.

Our mechanism occupies that gap: a weak label changes \emph{how peaked} the distillation target is at a given location, rather than which locations are shown or how their losses are scaled. DINO-LG is its view-level predecessor, and in this paper it also serves as the controlled baseline (Sec.~\ref{sec:backbones}), which lets us separate what target-level conditioning adds from what view-level guidance already provided.

\section{SALT: Self-Distillation with Lesion-Conditioned Targets}
\label{sec:pretrain}

\subsection{Background: self-distillation}
Our encoder is pretrained by self-distillation, following the DINOv2 recipe~\cite{dinov2}, which couples an image-level and a patch-level objective. In the image-level objective~\cite{dino} a \emph{student} and a \emph{teacher} network share the same architecture; the teacher is never updated by gradients but is an exponential moving average (EMA) of the student. From each image a \emph{multi-view} set is drawn by random resized cropping and photometric augmentation, two high-resolution \emph{global} views and several low-resolution \emph{local} views. Both networks embed a view and project its class token through a multi-layer head onto a $K$-dimensional distribution over learned prototypes. The teacher's distribution is made peaky by a low softmax temperature and stabilized by subtracting a running batch mean (\emph{centering}); these two operations together prevent the collapse to a constant output. Training minimizes the cross-entropy from the teacher (the target, with the gradient stopped) to the student over mismatched view pairs, so that a partial or corrupted view must reproduce the teacher's representation of a global view; the teacher is then nudged toward the student by the EMA step.

The patch-level objective~\cite{ibot} is a masked-image counterpart. The student additionally receives the global views with a random subset of their patch tokens replaced by a shared learned mask token, and must predict, at each masked position, the teacher's token distribution for that same patch computed from the \emph{intact} view. The teacher therefore acts as an online tokenizer that is learned jointly rather than fixed in advance. The image-level and patch-level objectives use separate projection heads, and a small regularizer (KoLeo) encourages the class embeddings within a batch to spread apart.

\subsection{Conditioning on weak lesion labels}
\begin{figure}[!htbp]
\centering
\includegraphics[width=\linewidth]{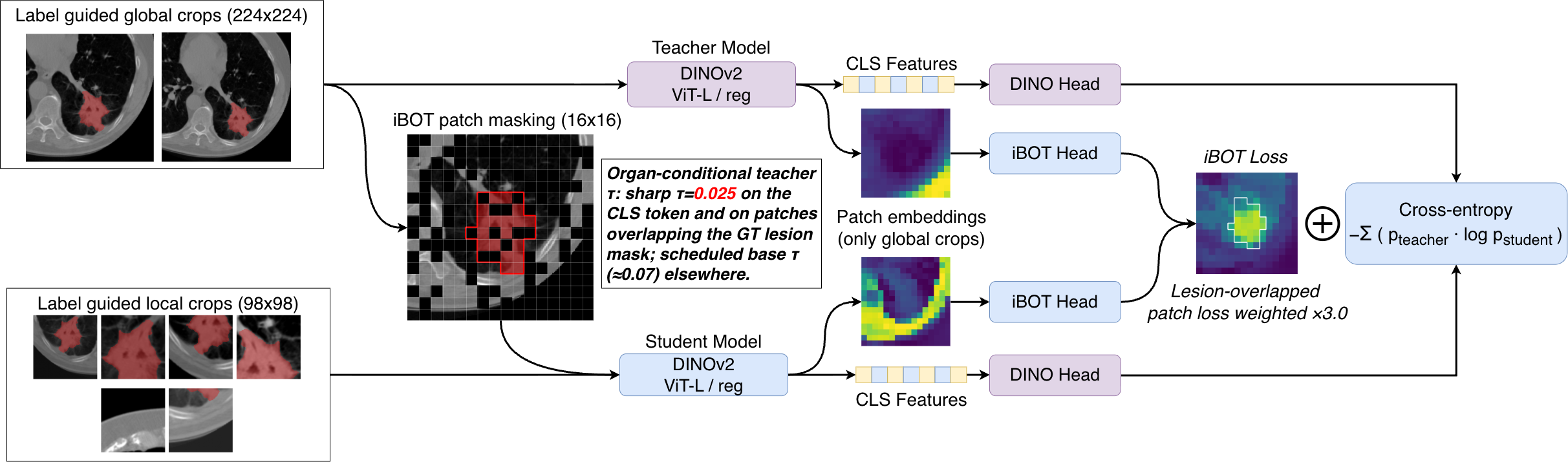}
\caption{Lesion-conditioned pretraining. The teacher encodes unmasked global views, the student masked globals
and local views; a sharper teacher temperature ($\tau_\ell\!=\!0.025$) is applied on the label-defined lesion
region.}
\label{fig:dino}
\end{figure}

Generic pretraining treats every spatial location alike: the teacher temperature and the loss weight are uniform across the image, so a lesion a few patches wide contributes no more to the objective than the surrounding parenchyma, and the encoder has little incentive to represent it distinctly. We therefore bias the pretraining \emph{targets} toward lesions using weak labels, bounding boxes converted to masks, that are available \emph{only during pretraining}; at every downstream use the encoder is a plain feature extractor with no labels and no conditioning. The objectives above are left unchanged; we add three label-driven mechanisms and preserve two properties that keep the targets well-behaved (Fig.~\ref{fig:dino}).

\textbf{Label-guided views.} For a labeled slice the two global views, and most of the local views, are cropped around a randomly chosen lesion pixel, so the lesion lies inside every conditioned view. The mask is subjected to the identical crop and horizontal flip as its view, so it stays pixel-aligned. Because the anchor pixel is redrawn each epoch, all lesions in a slice are covered over the course of training. This mechanism is view-level guidance as introduced in DINO-LG~\cite{dinolg}; it is shared by our conditioned encoders and by the DINO-LG baseline of Sec.~\ref{sec:backbones}, and is therefore \emph{not} what the comparison in this paper isolates. The two mechanisms that follow are.

\textbf{Compact lesion region.} Each view's mask is pooled onto the encoder's patch grid, marking a patch as lesion if it contains any lesion pixel. To avoid conditioning on incidental structures near the view border, only the connected group of lesion patches nearest the view center is kept (obtained by a bounded flood fill) and is then dilated by a few patches into a compact square region. This yields a stable footprint even when the lesion is smaller than a single patch.

\textbf{Region-conditioned teacher target.} Inside this region the teacher's softmax temperature is lowered,
\begin{equation}
\tau(p) \;=\; \tau_{\text{base}} - \big(\tau_{\text{base}} - \tau_{\ell}\big)\, w(p),
\qquad w(p)=\mathbf{1}[\,p \in \text{region}\,],
\end{equation}
with $\tau_{\ell}\!=\!0.025$ (sharp) and $\tau_{\text{base}}$ the standard warmup schedule ($0.04\!\rightarrow\!0.07$). A sharper teacher target where the lesion is pushes the student to commit to a more decisive code there. The same sharp temperature is applied to the class-token target of labeled views, so that whole-image summaries of lesion-bearing slices are likewise sharpened. The cut is hard, interior versus exterior, with the dilated region as the entire conditioned footprint. Finally, masked patches falling inside the region are up-weighted (by a factor of three) in the patch loss, so the objective attends more to lesion patches once they are masked.

\textbf{One mechanism, three parts.} The compact region exists solely to carry the sharpened temperature and the patch-loss weight, so the three components form a single conditioning mechanism and are introduced, and evaluated, as one.

\textbf{What is left untouched.} The conditioning enters only through the teacher's target temperature and the loss weighting; masking and every other component follow the standard recipe. Two properties then keep the conditioned targets calibrated: the teacher always encodes the \emph{intact} view when producing patch targets, so those targets reflect real content rather than the teacher's own inpainting of a hidden region; and the centering statistics are accumulated on the raw teacher outputs \emph{before} any temperature is applied, so the running mean that stabilizes training never sees the per-location sharpening.

\subsection{Configuration}
The encoder is a single-channel ViT-L with register tokens ($1024$-d, $24$ layers, patch size $14$), so a $224\times224$ view yields a $16\times16$ patch grid. We use two global views ($224^2$, area fraction $[0.4,1.0]$) and eight local views ($98^2$, $[0.1,0.4]$), six of the eight local views being lesion-guided. The two objectives use separate heads ($K\!=\!65536$); a random $10$--$50\%$ of patches is masked, the teacher momentum is $0.996$, and the student temperature is $0.1$. We optimize with AdamW for $75$k iterations at batch size $128$ (bf16, sharded), learning rate $10^{-4}\!\rightarrow\!10^{-5}$ (cosine), weight decay $0.04\!\rightarrow\!0.4$. Weak labels are drawn from an eight-cohort CT lesion store, restricted to training-split volumes. Table~\ref{tab:hparams} lists the full pretraining and detector-head configuration.

\begin{table}[!htbp]\centering\small
\begin{tabular}{ll|ll}
\hline
\multicolumn{2}{c|}{Pretraining (DINOv2/iBOT with SALT)} & \multicolumn{2}{c}{Detector head}\\
\hline
Encoder & ViT-L/14, 1024-d, 24L, reg. & Frozen taps & blocks $\{6,12,18,24\}$\\
Global views & 2 $\times 224^2$, scale $0.4$--$1.0$ & Trunk & stem $\to384$, 6 ResBlocks\\
Local views & 8 $\times 98^2$, scale $0.1$--$0.4$ (6 guided) & Dilations & $1,1,2,2,4,1$\\
Head out-dim $K$ & 65536 (both heads) & Input / grid & $512$ / $36\!\to\!128$\\
Mask prob / ratio & $0.5$ / $0.1$--$0.5$ & Center loss & focal $\alpha{=}2,\beta{=}4$\\
Teacher temp $\tau_{\mathrm{base}}$ & $0.04\!\to\!0.07$ (15k warmup) & Presence loss & BCE, $\lambda{=}1$\\
Lesion temp $\tau_\ell$ & $0.025$ (dilate $3$) & Box loss & L1 @centers, $\gamma{=}0.1$\\
Label loss weight & $\times 3.0$ & Match radius & $1.5$ (@grid $36$)\\
Teacher momentum & $0.996$ & Box pad & $+1$ cell (128-grid)\\
Student temp & $0.1$; center mom.\ $0.9$ & Presence gate & $0.5$ (train-time)\\
Optimizer & AdamW, clip $3.0$ & Epochs & $50$, AdamW cosine\\
LR & $10^{-4}\!\to\!10^{-5}$ cosine & Samples/epoch & 30000\\
Weight decay & $0.04\!\to\!0.4$ & Scope & lung $+$ lymph node\\
Iterations / batch & 75k / 128 &  & \\
Warmup (iter/temp) & 10k / 15k; freeze-last $300$ &  & \\
KoLeo weight & $0.1$ &  & \\
\hline\end{tabular}
\caption{Pretraining and detector-head hyperparameters. The two ViT-B variants differ from this configuration
only in encoder width/depth and, for the DINO-LG baseline, in the absence of the lesion-temperature and
label-loss-weight rows.}
\label{tab:hparams}
\end{table}

\section{Lesion Detection}
\label{sec:detect}

\subsection{Frozen encoder and multi-depth features}
The pretrained encoder is \emph{frozen}: it is used only to produce features and receives no gradients. A single forward pass exposes the token sequence after every transformer block; we take the patch tokens from four evenly spaced depths (blocks $6,12,18,24$ of the $24$-layer trunk) and concatenate them along the channel axis, giving a $36\times36$ grid of $4\times1024\!=\!4096$-dimensional descriptors for a $512\times512$ input. Reading several depths, rather than the final layer alone, combines mid-level texture with high-level semantics in a single descriptor.

\subsection{Detection head}
\begin{figure}[!htbp]
\centering
\includegraphics[width=\linewidth]{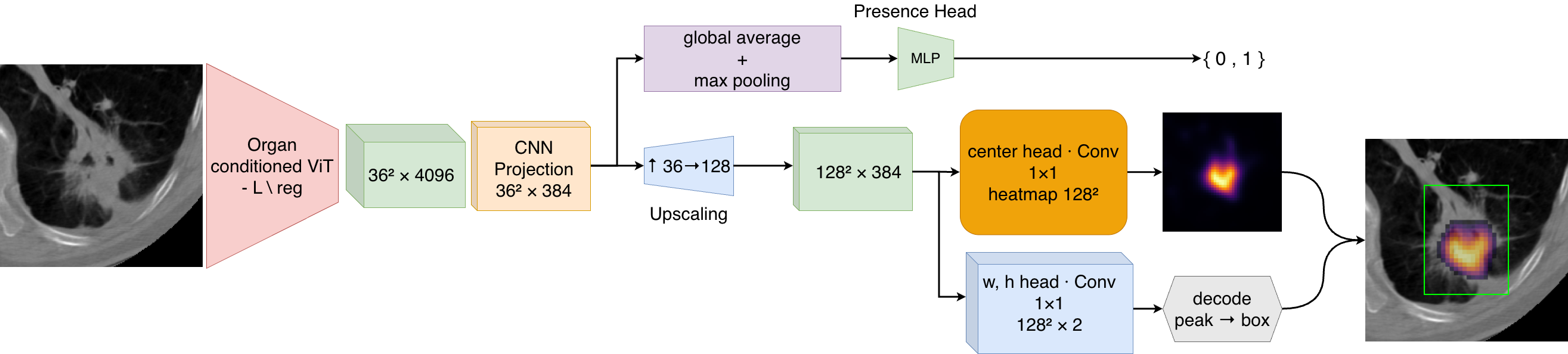}
\caption{Frozen-encoder detector. Multi-depth ViT patch tokens ($36^2\times4096$) are projected to a
$36^2\times384$ convolutional trunk. The slice-presence head pools this trunk globally and classifies it,
while the trunk is separately upsampled to $128^2$ and read by two $1\times1$ convolutions, a center-response
head and a box-size head, whose peaks are decoded into boxes.}
\label{fig:detector}
\end{figure}

Only a compact head is trained (Fig.~\ref{fig:detector}); one such head is trained per frozen encoder, and the encoder itself is never updated. A $3\times3$ stem projects the $4096$-channel multi-depth descriptor to $384$ channels, and six residual blocks with dilations $1,1,2,2,4,1$ enlarge the receptive field while holding the $36\times36$ resolution. This shared trunk is where the head branches, and it branches at two different resolutions.

The \emph{slice-level} head reads the trunk directly at $36\times36$: the feature map is reduced by global average and global max pooling, and the concatenated vector passes through a small multilayer perceptron that emits one logit for ``this slice contains a lesion''. Presence is a whole-slice question, so it needs no spatial refinement, and the head reads pooled convolutional features rather than the encoder's class token.

Localization instead needs finer spatial resolution than $36\times36$ affords, since one cell spans $14$ input pixels. The trunk is therefore upsampled to $128\times128$ by bilinear interpolation followed by two refinement convolutions, which sharpen where within a patch a center lies without fabricating detail the encoder never resolved. Two $1\times1$ convolutions then read this refined map, following the CenterNet keypoint-detection paradigm~\cite{centernet}: a \emph{center} head giving a single-channel response whose peaks mark lesion centers, and a \emph{box-size} head giving two channels, the width and height of the lesion box at each location. At inference the peaks of the center response are decoded into boxes by reading the box-size channels at the peak (Sec.~\ref{sec:detect}).

\subsection{Training targets and loss}
Regressing a unit bump at every $2$D centroid would ignore that a lesion is a $3$D object and that detection is ultimately scored in $3$D. We therefore first reassemble each lesion in $3$D by linking its per-slice cross-sections across adjacent slices, and set the target amplitude at a slice's centroid to the ratio of that slice's cross-sectional area to the lesion's \emph{largest} cross-sectional area over all its slices. The slice on which the lesion is largest becomes a hard center of value one, while thinner end-slices receive proportionally smaller targets, and the Gaussian width scales with $\sqrt{\text{area}}$ so that large and small lesions are represented commensurately. This concentrates confidence on the most informative slice of each lesion. The box-size target is the bounding box of the lesion mask, and the slice target records whether the slice contains any in-scope lesion.

The head is trained with
\begin{equation}
\mathcal{L} \;=\; \mathcal{L}_{\text{focal}}(\text{center})
\;+\; \lambda\,\mathrm{BCE}(\text{slice})
\;+\; \gamma\,\mathrm{L1}(\text{box}),
\end{equation}
a penalty-reduced focal loss on the center response ($\alpha\!=\!2,\beta\!=\!4$), a binary cross-entropy on the slice logit ($\lambda\!=\!1$), and an $\ell_1$ loss on the box size evaluated \emph{only} at true lesion centers ($\gamma\!=\!0.1$). We train for $50$ AdamW epochs (cosine schedule) on $512\times512$ inputs, restricted to lung and lymph-node lesions.

\subsection{Inference}
\begin{figure}[!htbp]
\centering
\includegraphics[width=\linewidth]{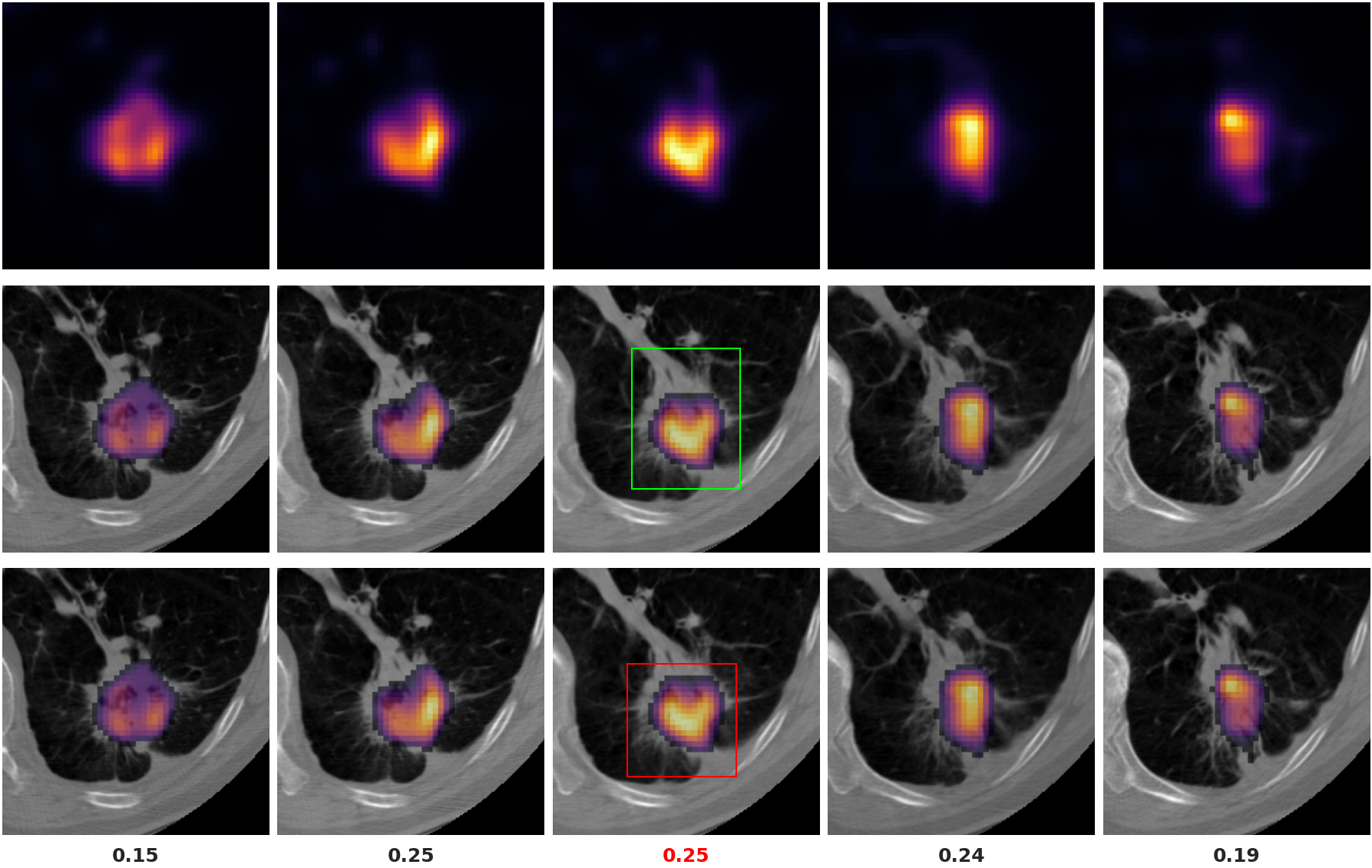}
\caption{Detector output over five consecutive slices of a lung lesion (NLST-seg). Rows: center-response
heatmap; CT$+$heatmap with the ground-truth box (green); CT$+$heatmap with the predicted box (red). Boxes on
the peak slice; max response below each panel.}
\label{fig:detout}
\end{figure}

Each slice is encoded and passed through the head. Candidate centers are the local maxima of the center response, identified by comparing each cell to its neighborhood via a pooling operation, that exceed a score threshold. For each candidate, a box is formed by reading the width and height at that location, centering them on the peak, converting grid cells to pixels, and padding by one cell. Candidate centers are then linked across adjacent slices by spatial proximity into $3$D detections, and the slice-level head acts as a gate. A $3$D lesion is reported as found when any of its slices is localized within a small tolerance of that slice's cross-section, the protocol under which we report results. Figure~\ref{fig:detout} shows the decoded center-response and boxes over five consecutive slices of a lung lesion.

\section{Experiments}
\label{sec:exp}

\subsection{Datasets and cohorts}
Our CT lesion store aggregates eight cohorts (Table~\ref{tab:cohorts}). Pretraining uses the training split of all eight; detection and re-identification are evaluated on the held-out test split of the in-scope cohorts. The test split is patient-disjoint from training (no patient appears in both), and lesion counts, volumes, patients and the split are read from the store metadata.

\begin{table}[!htbp]\centering\small\setlength{\tabcolsep}{3.5pt}
\begin{tabular}{lllllrrrrr}
\hline
Cohort & Anatomy & Annot. & Access & Usage & \#Pat & \#Vol & \#Slice & tr/te$_{\mathrm{vol}}$ & \#Les$_{\mathrm{test}}$\\
\hline
DeepLesion~\cite{deeplesion} & multi-organ & box & public & pretrain (weak) & 4427 & 14601 & 928k & 14601/0 & --\\
UniToChest~\cite{unitochest} & lung & mask & public & pretrain $+$ eval & 623 & 715 & 306k & 546/169 & 3008\\
NLST-seg~\cite{nlst} & lung & mask & public & pretrain $+$ eval & 604 & 604 & 107k & 481/123 & 149\\
LiTS~\cite{lits} & liver & mask & public & pretrain (o.o.s.) & 131 & 131 & 59k & 101/30 & --\\
NIH-Lymph~\cite{nihlymph} & lymph node & mask & public & pretrain $+$ eval & 89 & 89 & 55k & 71/18 & 285\\
HCUCH~\cite{hcuch} & mixed & mask & public & pretrain $+$ eval $+$ re-id & 58 & 58 & 14k & 38/20 & 541\\
T\"ubingen~\cite{tuebingen} & lung/node/liver & mask & public & pretrain & 300 & 670 & 200k & 494/176 & --\\
LyNoS~\cite{lynos} & lymph node & mask & public & pretrain & 15 & 15 & 9k & 13/2 & --\\
\hline
Total & -- & -- & -- & -- & 6247 & 16883 & 1679k & -- & 3983\\
\hline\end{tabular}
\caption{Cohorts of the $8$-source CT lesion store. Annotation: DeepLesion is RECIST-\emph{box}, the rest voxel
\emph{mask}. \#Pat/\#Vol/\#Slice and the volume-level train/test split are read from the store metadata; the eval
cohorts' test split is patient-disjoint. \#Les$_{\mathrm{test}}$ is the $3$D lesion count used in detection
(Table~\ref{tab:detcoh}); HCUCH re-identification uses a $28$-lesion lung/lymph RECIST subset. o.o.s.\ = liver,
outside the lung$+$lymph detection scope.}
\label{tab:cohorts}
\end{table}

\subsection{Backbones compared}
\label{sec:backbones}
We compare our encoders against four external frozen backbones spanning three pretraining regimes (Table~\ref{tab:backbones}). Because our task is lesion localization, the salient axis is the pretraining \emph{data}: our encoders are the only ones pretrained on lesion-labeled CT, whereas the baselines see general CT, mixed CT/MR radiology, or natural images. \emph{Ours} are single-channel DINOv2-with-registers encoders pretrained on our CT lesion store (training split only): a ViT-L/14 with lesion-conditioned targets (Sec.~\ref{sec:pretrain}, our main model) and two ViT-B/14 variants, one with the same conditioning, and one with \emph{label-guided views only}, i.e.\ the DINO-LG recipe~\cite{dinolg} carried over to DINOv2. The latter is the controlled baseline of this paper: identical architecture, patch size, pretraining data, optimizer, schedule, seed and lesion-guided cropping, differing from the SALT ViT-B \emph{only} in the target-level mechanisms of Sec.~\ref{sec:pretrain} (the compact region, the sharpened teacher temperature and the up-weighted patch loss). It therefore isolates target-level conditioning from view-level guidance, and the ViT-L isolates scale. \textbf{MedDINOv3}~\cite{meddinov3} is a ViT-B/16 pretrained with a three-stage DINOv3~\cite{dinov3} recipe on CT-3M ($\approx\!3.9$M axial CT slices from $16$ public datasets), for CT segmentation. \textbf{Curia}~\cite{curia} and \textbf{Curia-2}~\cite{curia2} are ViT-B/16 DINOv2 radiology foundation models (Raidium) pretrained on $\approx\!200$M CT and MR $2$D slices from $\approx\!150$k private hospital exams. \textbf{DINOv2 with registers}~\cite{dinov2,registers} is Meta's ViT-L/14 pretrained on $142$M natural images (no medical data), a domain-transfer reference. Across all pools only \emph{LiTS} (liver tumour) is shared with ours, it also seeds MedDINOv3's CT-3M, and even there MedDINOv3 pretrains without labels, so no lesion \emph{annotation} overlaps; Curia's corpus is private and DINOv2's is natural. All backbones are frozen; a $512$ input yields a $36\times36$ patch grid at patch-$14$ and $32\times32$ at patch-$16$.

\begin{table}[!htbp]
\centering
\small
\setlength{\tabcolsep}{4pt}
\renewcommand{\arraystretch}{1.15}
\begin{tabular}{lllccrcp{4.0cm}}
\hline
Backbone & Method & Arch & Patch & Feat & \#Slices & \shortstack{Pretrain\\labels} & Pretraining datasets\\
\hline
SALT ViT-L            & DINOv2 $+$ SALT & ViT-L & /14 & 1024 & $1.5$M & \textbf{target} & DeepLesion, UniToChest, NLST, LiTS, NIH-Lymph, HCUCH, T\"ubingen, LyNoS, CT, weak box$\to$mask labels (ours)\\[2pt]
SALT ViT-B            & DINOv2 $+$ SALT & ViT-B & /14 & 768  & $1.5$M & \textbf{target} & (same $8$-cohort store)\\[2pt]
DINO-LG ViT-B                 & DINO-LG, view-guided & ViT-B & /14 & 768  & $1.5$M & views only & (same store; label-guided views, no target conditioning)\\[2pt]
MedDINOv3~\cite{meddinov3}           & DINOv3               & ViT-B & /16 & 768  & $3.9$M & no & CT-3M: $16$ public CT sets, incl.\ LiTS~\cite{lits}, KiTS~\cite{kits}, AMOS22~\cite{amos}, WORD~\cite{word}, BTCV~\cite{btcv}, CHAOS~\cite{chaos}, AbdomenCT-1K~\cite{abdomenct1k}, MSD~\cite{msd}, CT-ORG~\cite{ctorg}, TotalSegmentator~\cite{totalsegmentator}, AbdomenAtlas~\cite{abdomenatlas}, Pancreas-CT~\cite{pancreasct}; natural init (LVD-1689M~\cite{dinov3})\\[2pt]
Curia~\cite{curia}                   & DINOv2               & ViT-B & /16 & 768  & $\sim$200M & no & Private hospital corpus, $\sim$150k CT$+$MR exams (datasets undisclosed)\\[2pt]
Curia-2~\cite{curia2}                & DINOv2               & ViT-B & /16 & 768  & undisc.   & no & Private, CT$+$MR (undisclosed)\\[2pt]
DINOv2$+$reg~\cite{dinov2,registers} & DINOv2$+$reg.        & ViT-L & /14 & 1024 & $142$M    & no & LVD-142M natural images (ImageNet-22k~\cite{imagenet}, Google Landmarks~\cite{googlelandmarks}, \dots)\\
\hline
\end{tabular}
\caption{Frozen backbones compared: architecture, patch size, feature dimension, pretraining slices, how lesion
labels enter pretraining, and the pretraining datasets by name. \#Slices: $2$D axial CT slices seen in
pretraining (natural images for DINOv2$+$reg.). ``Pretrain labels'' distinguishes \emph{target} conditioning
(ours, Sec.~\ref{sec:pretrain}) from \emph{view}-level guidance alone (DINO-LG~\cite{dinolg}) and from no label
use at all; the DINO-LG row is our controlled baseline. Patient counts are not shown for the external models:
their sources report exams/slices, not patients. Parameter counts are $304$M for ViT-L and $86$M for ViT-B.}
\label{tab:backbones}
\end{table}

\subsection{Lesion feature discriminability}
Before attaching any detector we ask how strongly each frozen backbone already separates lesion tissue from its surroundings, using its patch embeddings alone (Table~\ref{tab:sep}).

\begin{table}[!htbp]
\centering
\small
\setlength{\tabcolsep}{6pt}
\begin{tabular}{lrrr}
\hline
Backbone & AUROC & sep & bg-sim\\
\hline
SALT ViT-L             & 0.995 & \textbf{0.517} & \textbf{0.165}\\
SALT ViT-B             & 0.995 & 0.449 & 0.239\\
DINOv2$+$reg (natural) & 0.993 & 0.250 & 0.566\\
Curia                  & 0.995 & 0.233 & 0.592\\
MedDINOv3              & 0.998 & 0.232 & 0.698\\
DINO-LG ViT-B          & 0.995 & 0.105 & 0.502\\
Curia-2                & 0.996 & 0.047 & 0.927\\
\hline
\end{tabular}
\caption{Lesion-vs-background patch separability of frozen backbones (one random volume per cohort, averaged
over four). Architectures and pretraining corpora are given in Table~\ref{tab:backbones}. AUROC: $5$-fold
cross-validated linear probe; sep: mean lesion$-$background cosine gap to a lesion prototype; bg-sim:
background similarity (lower is better). Protocol and metric definitions below.}
\label{tab:sep}
\end{table}

\textbf{How it is measured.} We draw one random volume from each of four cohorts (UniToChest, NLST-seg, NIH-Lymph, HCUCH) at full resolution, the \emph{same} four volumes for every backbone, and encode every slice with the frozen backbone. Each patch is labeled lesion or background by pooling the segmentation mask onto the patch grid. We then report two quantities, averaged over the four volumes. (i) A scale-invariant \emph{linear probe}: a logistic regression is fit to separate lesion from background patch embeddings and scored by $5$-fold cross-validated AUROC; it asks only whether a lesion direction \emph{exists}, not how prominent it is. (ii) The \emph{cosine geometry} of the raw embeddings: we form a lesion prototype (the mean lesion-patch embedding), score every patch by cosine similarity to it, and report $\mathrm{sep}=\text{mean}_{\text{lesion}}- \text{mean}_{\text{bg}}$ of that similarity (the separation margin; higher is better) and $\text{bg-sim}=\text{mean}_{\text{bg}}$ (how strongly non-lesion tissue also responds; lower is better).

\textbf{Result.} The linear probe is near-perfect for every backbone (AUROC column, all $\ge0.99$): a lesion-versus-background direction is linearly present in all of them, medical or natural, small or large. The backbones diverge sharply, however, in how \emph{geometrically concentrated} that signal is, and the two measures can even point in opposite directions, MedDINOv3 has the highest probe AUROC ($0.998$) and the second-worst background similarity ($0.698$), so linear decodability is no guarantee of a concentrated representation. Our two SALT encoders are the top two on the geometry: the ViT-L reaches a separation margin of $0.52$ and the ViT-B $0.45$, both well ahead of the strongest baseline ($0.25$), with the lowest background similarities ($0.17$ and $0.24$).

\begin{figure}[!htbp]
\centering
\includegraphics[width=0.31\linewidth]{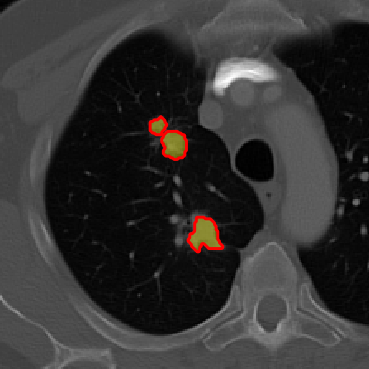}\hfill
\includegraphics[width=0.31\linewidth]{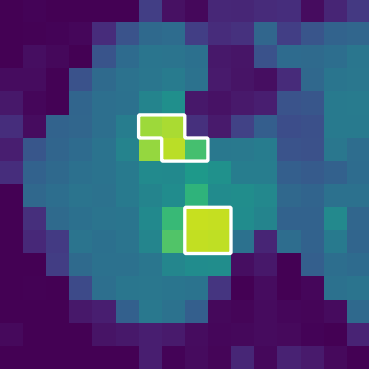}\hfill
\includegraphics[width=0.31\linewidth]{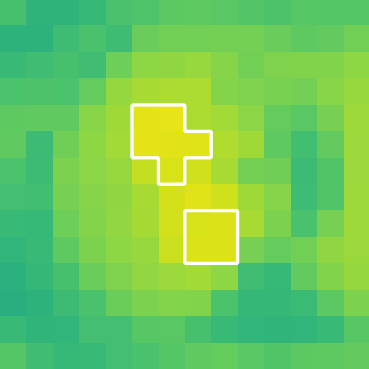}
\caption{Per-patch cosine similarity to the lesion prototype on one UniToChest slice (lesion outline white).
Left to right: CT, our SALT ViT-L, MedDINOv3.}
\label{fig:signal}
\end{figure}

On this representation-level measure the \emph{target-level} conditioning is decisive, and the matched pair isolates it: the DINO-LG ViT-B, identical in architecture, pretraining data, schedule and label-guided cropping, and differing only in the target conditioning, reaches only $0.105$, a $\sim\!4\times$ gap at ViT-B and the lowest margin of any backbone except Curia-2. Two readings follow. First, sharpening the teacher target does what it was designed to do: lesion patches form a tight, well-separated cluster while unrelated tissue is pushed away. Second, view-level guidance \emph{on its own} does not produce this geometry, and here leaves it below both generic CT (MedDINOv3, Curia) and the natural-image reference, consistent with the view that restricting crops narrows the training distribution without giving the objective any reason to represent the lesion distinctly. Figure~\ref{fig:signal} visualizes this score: each patch of one slice is colored by its cosine similarity to the lesion prototype (the quantity whose in-lesion/background gap is sep and whose background mean is bg-sim) and upsampled to the image grid, for our ViT-L the lesion lights up while the background stays dark, whereas the generic MedDINOv3 responds broadly.

\subsection{Detection}
Attaching the frozen-backbone detector of Sec.~\ref{sec:detect} to each backbone, we evaluate $3$D catch-once detection on the four cohorts ($3983$ lesions). A \emph{separate} head is trained for each backbone, since the descriptor width depends on the encoder ($4\times1024$ channels for ViT-L, $4\times768$ for ViT-B) and the patch grid on the patch size ($36\times36$ at patch-$14$, $32\times32$ at patch-$16$). These two dimensions are fixed by the frozen encoder and are the only respect in which the runs differ: the stem's input width follows the descriptor, so the ViT-B heads carry a slightly smaller stem, while everything downstream of it is unchanged. Every other choice, the trunk width and depth, the dilations, the loss, the optimizer, the schedule and the random seed, is identical in every run, and no encoder receives gradients. Differences between rows therefore reflect the frozen features alone (Table~\ref{tab:det}).

\begin{table}[!htbp]
\centering
\small
\setlength{\tabcolsep}{4.5pt}
\begin{tabular}{lrrrrrrr}
\hline
Backbone & R & P & $F_1$ & R$_{\max}$ & FP/sc & CPM & thr\\
\hline
SALT ViT-L              & 0.611 & 0.783 & \textbf{0.686} & 0.785 & 2.05 & \textbf{0.556} & 0.20\\
DINOv2$+$reg (natural)  & 0.592 & 0.767 & 0.668 & 0.811 & 2.18 & 0.518 & 0.35\\
SALT ViT-B              & 0.554 & 0.795 & 0.653 & 0.751 & 1.74 & 0.508 & 0.20\\
Curia                   & 0.492 & 0.729 & 0.588 & 0.744 & 2.22 & 0.433 & 0.25\\
MedDINOv3               & 0.485 & 0.679 & 0.566 & 0.730 & 2.79 & 0.423 & 0.20\\
DINO-LG ViT-B           & 0.457 & 0.709 & 0.556 & 0.728 & 2.28 & 0.394 & 0.25\\
Curia-2                 & 0.383 & 0.617 & 0.473 & 0.658 & 2.89 & 0.315 & 0.25\\
\hline
\end{tabular}
\caption{$3$D catch-once detection over the four cohorts ($3983$ lesions). Each encoder is frozen and carries
its own head, trained with the same architecture, data, loss and schedule. R/P/$F_1$ at the best-$F_1$
threshold (thr, per model); R$_{\max}$ the recall ceiling (lesion-weighted mean of the per-cohort ceilings);
FP/sc false positives per scan; CPM the threshold-free mean FROC sensitivity ($0.5$--$4$ FP/scan).
Slice-presence performance is reported separately in Table~\ref{tab:pres}. Wilson $95\%$ CI half-widths are
$\le0.016$ on R ($n{=}3983$ lesions) and $\le0.019$ on P. Protocol and metric definitions below.}
\label{tab:det}
\end{table}

\textbf{How it is measured.} Each test slice is encoded and scored by the head; per-slice peak detections are linked across adjacent slices into $3$D candidates, and a $3$D ground-truth lesion counts as detected (\emph{catch-once}) if \emph{any} single one of its slices is localized within a small tolerance of that slice's cross-section. From the resulting $3$D true/false detections we report, at the score threshold that maximizes $F_1$: recall (R), precision (P), $F_1$, and false positives per scan (FP/sc). This detection threshold, and the presence-gate threshold (Table~\ref{tab:pres}), maximizes $F_1$ \emph{on the test set itself}; we hold out no separate validation split, so R/P/$F_1$, and every quantity read at the operating point, are best-case figures.

\paragraph{CPM.} Any single operating point is an arbitrary choice, and two detectors can trade recall against false positives so differently that comparing them at one threshold is misleading. The competition performance metric (CPM), introduced with the LUNA16 lung-nodule challenge~\cite{luna16}, avoids this by summarizing the whole operating curve. Concretely, we sweep the detection score threshold from high to low. At each value we count how many ground-truth $3$D lesions are found (sensitivity) and how many false detections are produced per scan; plotting sensitivity against false positives per scan traces the free-response ROC (FROC) curve, the detection analogue of an ROC curve in which the horizontal axis is an unbounded count of false alarms per case rather than a false-positive rate. Lowering the threshold moves rightward along this curve, buying sensitivity at the cost of false positives. CPM is then the mean sensitivity at a fixed set of false-positive rates,
\begin{equation}
\mathrm{CPM} \;=\; \frac{1}{|\mathcal{F}|}\sum_{f \in \mathcal{F}} \mathrm{Sens}(f),
\qquad \mathcal{F} = \{0.5,\,1,\,2,\,4\}\ \text{FP/scan},
\end{equation}
where $\mathrm{Sens}(f)$ is the recall at the threshold whose false-positive rate is $f$, obtained by interpolation on the FROC curve. A single number therefore reports how much of the lesion population a detector recovers across a clinically plausible range of false-alarm budgets, from a strict false-alarm budget ($0.5$ FP/scan) to a permissive one ($4$ FP/scan). Because no threshold is selected, CPM is unaffected by the test-set threshold choice above; it is our \emph{primary} detection metric, as are R$_{\max}$ and every AUROC reported here.

We evaluate on $\mathcal{F}=\{0.5,1,2,4\}$ FP/scan, whereas the LUNA16 convention averages over $\{0.125,0.25,0.5,1,2,4,8\}$. Our range omits the two strictest and the loosest budgets, so the values here are not directly comparable with CPM figures reported on LUNA16, and we make no such comparison.

All figures pool the four cohorts.

\textbf{Result.} Our SALT ViT-L is the strongest detector (CPM $=\!0.556$, $F_1\!=\!0.686$). Pooled over the four cohorts, target-level conditioning is worth $+0.11$ CPM at matched scale: against the DINO-LG ViT-B, same architecture, data, schedule and label-guided cropping, it lifts CPM from $0.394$ to $0.508$ and $F_1$ from $0.556$ to $0.653$, at a lower false-positive rate ($1.74$ vs $2.28$ per scan) and a higher recall ceiling ($0.751$ vs $0.728$). Because the operating threshold is selected the same way for both, the $F_1$ difference compares two upper bounds and is indicative; the CPM difference carries the claim. This pooled figure, however, averages a \emph{heterogeneous} effect: Sec.~\ref{sec:cohortres} shows it is large on two cohorts, slightly negative on the other two, and dominated by the cohort that supplies three quarters of the lesions. Both SALT encoders outrank the dedicated CT/MR foundation models Curia (CPM $0.433$) and MedDINOv3 ($0.423$); the only baseline that keeps pace is the natural-image DINOv2 ViT-L ($0.518$), which our ViT-L tops at equal scale and equal patch size, so scale helps detection but does not substitute for lesion-aware pretraining.

\subsection{Per-cohort behavior}
\label{sec:cohortres}
Table~\ref{tab:detcoh} breaks detection down by cohort for every backbone (same catch-once protocol and metrics as Table~\ref{tab:det}); the size and box analyses that follow focus on our SALT ViT-L.

\begin{table}[!htbp]
\centering
\small
\setlength{\tabcolsep}{4pt}
\begin{tabular}{llrrrrr}
\hline
Cohort & Backbone & R & P & $F_1$ & R$_{\max}$ & CPM\\
\hline
NLST-seg (lung, $n{=}149$) & SALT ViT-L  & 0.705 & 0.556 & 0.621 & 0.980 & 0.628\\
 & SALT ViT-B  & 0.651 & 0.678 & 0.664 & 0.940 & 0.565\\
 & DINO-LG ViT-B       & 0.711 & 0.527 & 0.606 & 0.973 & 0.633\\
 & MedDINOv3                  & 0.738 & 0.482 & 0.584 & 0.987 & 0.621\\
 & Curia                      & 0.705 & 0.510 & 0.592 & 0.973 & 0.616\\
 & Curia-2                    & 0.718 & 0.382 & 0.499 & 0.973 & 0.601\\
 & DINOv2$+$reg (natural)     & 0.752 & 0.505 & 0.604 & 0.993 & 0.634\\
\hline
UniToChest (lung, $n{=}3008$) & SALT ViT-L  & 0.600 & 0.782 & 0.679 & 0.789 & 0.544\\
 & SALT ViT-B  & 0.533 & 0.784 & 0.635 & 0.753 & 0.495\\
 & DINO-LG ViT-B       & 0.409 & 0.698 & 0.516 & 0.716 & 0.351\\
 & MedDINOv3                  & 0.439 & 0.659 & 0.527 & 0.723 & 0.372\\
 & Curia                      & 0.464 & 0.722 & 0.565 & 0.742 & 0.410\\
 & Curia-2                    & 0.337 & 0.597 & 0.431 & 0.633 & 0.283\\
 & DINOv2$+$reg (natural)     & 0.565 & 0.786 & 0.658 & 0.812 & 0.492\\
\hline
NIH-Lymph (node, $n{=}285$) & SALT ViT-L  & 0.639 & 0.858 & 0.732 & 0.898 & 0.555\\
 & SALT ViT-B  & 0.733 & 0.833 & 0.780 & 0.902 & 0.656\\
 & DINO-LG ViT-B       & 0.533 & 0.752 & 0.624 & 0.863 & 0.446\\
 & MedDINOv3                  & 0.558 & 0.723 & 0.630 & 0.839 & 0.496\\
 & Curia                      & 0.425 & 0.720 & 0.534 & 0.825 & 0.388\\
 & Curia-2                    & 0.418 & 0.773 & 0.542 & 0.818 & 0.285\\
 & DINOv2$+$reg (natural)     & 0.611 & 0.688 & 0.647 & 0.916 & 0.569\\
\hline
HCUCH (mixed, $n{=}541$) & SALT ViT-L  & 0.630 & 0.859 & 0.727 & 0.647 & 0.602\\
 & SALT ViT-B  & 0.545 & 0.883 & 0.674 & 0.612 & 0.486\\
 & DINO-LG ViT-B       & 0.614 & 0.824 & 0.703 & 0.652 & 0.540\\
 & MedDINOv3                  & 0.634 & 0.866 & 0.732 & 0.641 & 0.610\\
 & Curia                      & 0.628 & 0.883 & 0.734 & 0.649 & 0.532\\
 & Curia-2                    & 0.529 & 0.841 & 0.649 & 0.628 & 0.431\\
 & DINOv2$+$reg (natural)     & 0.686 & 0.851 & 0.759 & 0.695 & 0.607\\
\hline
\end{tabular}
\caption{Per-cohort detection for \emph{all} backbones. R/P/$F_1$ at each model's operating point; R$_{\max}$ the
recall ceiling at the loosest FP tolerance ($8$ FP/scan); CPM the mean FROC sensitivity (as in
Table~\ref{tab:det}). \#lesions per cohort in the header. UniToChest supplies $75.5\%$ of the pooled lesions, so
the pooled figures of Table~\ref{tab:det} are close to a UniToChest average.}
\label{tab:detcoh}
\end{table}

\textbf{Result.} The four cohorts differ sharply in how far apart they place the backbones. On NLST-seg all seven encoders fall within $0.069$ CPM of one another, and on HCUCH within $0.179$, whereas UniToChest spreads them over $0.261$ and NIH-Lymph over $0.371$. Two of the cohorts therefore carry almost no discriminative information about the encoders, and any ranking read from them is close to arbitrary.

Consistent with this, the ordering of backbones is not stable across cohorts: the best encoder is the natural-image DINOv2 on NLST-seg, our conditioned ViT-L on UniToChest, our SALT ViT-B on NIH-Lymph and MedDINOv3 on HCUCH. The pooled ordering of Table~\ref{tab:det} reproduces the UniToChest ordering almost exactly, which is what one expects given that this cohort supplies $75.5\%$ of the evaluation lesions. What does hold everywhere is that one of our conditioned encoders is never worse than third of seven, and is first on both cohorts that separate the field.

\paragraph{The conditioning gain is cohort-dependent.} Reading the matched ViT-B pair down the rows of Table~\ref{tab:detcoh}, the effect is plainly not uniform. Conditioning raises CPM substantially on UniToChest ($0.351\!\rightarrow\!0.495$, $+0.144$) and NIH-Lymph ($0.446\!\rightarrow\!0.656$, $+0.210$), and lowers it on HCUCH ($0.540\!\rightarrow\!0.486$, $-0.054$) and NLST-seg ($0.633\!\rightarrow\!0.565$, $-0.068$). The pooled $+0.114$ of Table~\ref{tab:det} is exactly the lesion-weighted mean of these four numbers, and because UniToChest alone supplies $3008$ of $3983$ lesions the pooled figure is close to the UniToChest figure: over the remaining $975$ lesions the residual gain is $+0.021$ CPM. \emph{The pooled improvement is therefore carried by two cohorts, one of which dominates the lesion count, and should not be read as a uniform benefit.}

The two cohorts without a gain are the two smallest contributors ($149$ and $541$ lesions), where a difference of this size amounts to roughly ten lesions, and they are the two that separate the encoders least. The largest single gain, by contrast, is on NIH-Lymph, the only pure lymph-node cohort, so the effect is not confined to one anatomy. With one pretraining seed and no interval on the per-cohort differences we cannot say whether the two negative signs reflect noise, a ceiling, or a real cost of conditioning on those distributions. What the matched pair does establish across every cohort is the representation-level effect of Table~\ref{tab:sep}, and, in the pooled and size-stratified views, a detection gain concentrated on the smallest lesions.

\subsection{Detection by lesion size}
Table~\ref{tab:detsize} breaks the same detector down by lesion diameter, and Table~\ref{tab:vitb_size} applies the same stratification to the matched ViT-B pair, the axis along which the conditioning gain concentrates.

\begin{table}[!htbp]
\centering
\small
\begin{tabular}{lrrrrr}
\hline
Lesion size & \#Les & \% & R & R$_{\max}$ & CPM\\
\hline
$<6$\,mm      & 3251 & 81.6 & 0.592 & 0.774 & 0.537\\
$6$--$10$\,mm  & 354  & 8.9  & 0.630 & 0.754 & 0.593\\
$10$--$20$\,mm & 305  & 7.7  & 0.741 & 0.892 & 0.677\\
$20$--$30$\,mm & 47   & 1.2  & 0.787 & 0.979 & 0.702\\
$>30$\,mm     & 26   & 0.7  & 0.808 & 1.000 & 0.760\\
\hline
\end{tabular}
\caption{Detection of our SALT ViT-L by lesion size (long-axis diameter). \% is the size bin's
share of all test lesions. Columns otherwise as in Table~\ref{tab:detcoh}. Most lesions are sub-$6$\,mm; recall
and CPM rise monotonically with size. Precision/$F_1$ are not broken out by size: a false positive has no
ground-truth size, so precision is undefined per size bin.}
\label{tab:detsize}
\end{table}

\begin{table}[!htbp]\centering\small
\begin{tabular}{lrrrrrr}
\hline
 & & \multicolumn{2}{c}{DINO-LG (view-guided)} & \multicolumn{2}{c}{SALT} & \\
Lesion size & \#Les & R & CPM & R & CPM & $\Delta$CPM\\
\hline
$<6$\,mm      & 3251 & 0.409 & 0.350 & 0.525 & 0.486 & $+0.136$\\
$6$--$10$\,mm  & 354  & 0.588 & 0.508 & 0.599 & 0.520 & $+0.012$\\
$10$--$20$\,mm & 305  & 0.748 & 0.680 & 0.754 & 0.681 & $+0.001$\\
$20$--$30$\,mm & 47   & 0.702 & 0.622 & 0.787 & 0.691 & $+0.069$\\
$>30$\,mm     & 26   & 0.769 & 0.654 & 0.769 & 0.740 & $+0.086$\\
\hline\end{tabular}
\caption{The matched ViT-B pair by lesion size: identical architecture, pretraining data, schedule and
label-guided cropping, differing only in the target-level conditioning. The gain is concentrated on sub-$6$\,mm
lesions ($+0.136$ CPM, $+0.116$ recall), which are $81.6\%$ of the test set, and vanishes between $6$ and
$20$\,mm. The two largest bins hold $47$ and $26$ lesions, too few to read as a separate effect. Precision/$F_1$
are undefined per size bin (Table~\ref{tab:detsize}); recall ceilings in the sub-$6$\,mm bin are $0.702$ and
$0.737$.}
\label{tab:vitb_size}
\end{table}

\textbf{Result.} Detection scales with lesion size: sub-$6$\,mm lesions, $81.6\%$ of the test set, are the hardest (CPM $0.537$), rising monotonically to $0.760$ above $30$\,mm. The matched pair shows that the conditioning gain is not spread across this range but concentrated at its lower end: on sub-$6$\,mm lesions the SALT ViT-B lifts CPM from $0.350$ to $0.486$ ($+0.136$) and recall from $0.409$ to $0.525$ ($+0.116$), while between $6$ and $20$\,mm the two encoders are indistinguishable ($+0.012$ and $+0.001$ CPM). Because the sub-$6$\,mm bin holds $3251$ of $3983$ lesions, it accounts for essentially all of the pooled difference in Table~\ref{tab:det}. This is the pattern the mechanism predicts: where a lesion spans many patches a spatially uniform objective already gives it substantial weight, whereas where it spans two or three a sharpened teacher target supplies the only local pressure to encode it distinctly.

\paragraph{Box quality.} We assess how well the regressed box recovers a lesion's spatial extent, in two read modes (Table~\ref{tab:box}).

\begin{table}[!htbp]
\centering
\small
\setlength{\tabcolsep}{5pt}
\begin{tabular}{lrr|rrr}
\hline
 & \multicolumn{2}{c|}{@GT center} & \multicolumn{3}{c}{@detected peak}\\
Backbone & cov & IoBB & cov & IoBB & det\%\\
\hline
SALT ViT-L  & 0.991 & 0.983 & \textbf{0.751} & \textbf{0.728} & 0.893\\
DINOv2$+$reg (natural)     & 0.960 & 0.933 & 0.647 & 0.621 & 0.903\\
SALT ViT-B  & 0.986 & 0.975 & 0.713 & 0.689 & 0.875\\
Curia                      & 0.987 & 0.976 & 0.707 & 0.684 & 0.854\\
MedDINOv3                  & 0.984 & 0.972 & 0.703 & 0.678 & 0.849\\
DINO-LG ViT-B       & 0.986 & 0.975 & 0.724 & 0.698 & 0.855\\
Curia-2                    & 0.985 & 0.974 & 0.722 & 0.701 & 0.777\\
\hline
\end{tabular}
\caption{Box quality across backbones: box-vs-mask coverage (cov) and box-vs-box IoBB, read at the GT lesion
center (box head isolated) and chained at the detected peak; det\% is the localization rate. Definitions and
result below.}
\label{tab:box}
\end{table}

\begin{table}[!htbp]\centering\small
\begin{tabular}{lrrrrr}
\hline
 & & \multicolumn{2}{c}{@GT center} & \multicolumn{2}{c}{@detected peak}\\
Size & \#Les & cov & IoBB & cov & IoBB\\
\hline
$<6$\,mm     & 3251 & 0.994 & 0.988 & 0.726 & 0.707\\
$6$--$10$\,mm  & 354 & 0.997 & 0.991 & 0.862 & 0.840\\
$10$--$20$\,mm & 305 & 0.969 & 0.944 & 0.862 & 0.820\\
$20$--$30$\,mm & 47  & 0.938 & 0.893 & 0.868 & 0.807\\
$>30$\,mm     & 26  & 0.869 & 0.799 & 0.741 & 0.646\\
\hline\end{tabular}
\caption{Box quality by lesion size (ViT-L). cov: box-vs-mask coverage; IoBB: box-vs-box overlap. @GT-center
isolates the box head; @detected-peak is the chained pipeline.}
\label{tab:box_size}
\end{table}

\textbf{How it is measured.} The box head is read in two modes. \emph{@GT center}: the box is read at the ground-truth lesion center, isolating box-regression quality from whether the detector fired. \emph{@detected peak}: the box is read at the nearest detected center peak, the real inference pipeline, and exists only when the lesion is localized. Two overlap metrics are reported: cov (coverage), the fraction of the ground-truth lesion \emph{mask} area enclosed by the predicted box; and IoBB, the intersection-over-union of the predicted and ground-truth \emph{boxes}. det\% is the localization rate (fraction of lesions with a detected peak near the GT center), which bounds how often the @detected-peak columns are defined.

\textbf{Result.} In isolation (@GT center) the box head is near-exact for \emph{every} backbone, covering $96$--$99\%$ of the lesion mask (IoBB $\ge0.93$), so recovering a lesion's spatial extent from a frozen patch grid is easy once its center is known, and the choice of backbone changes little (only the natural-image DINOv2 trails, at $0.96$ coverage). Chained (@detected peak), box quality instead tracks \emph{localization}: our SALT ViT-L gives the best real-pipeline coverage ($0.751$ at an $89\%$ localization rate), The natural-image DINOv2 is the exception: it localizes the most lesions ($90\%$, in line with its high recall) yet its boxes cover the least ($0.647$), i.e.\ it \emph{finds} lesions well but \emph{delineates} them less accurately. The gap between the two modes is thus driven by missed or soft peaks, not by box-regression error, and it is smallest for our ViT-L. Stratified by size (Table~\ref{tab:box_size}), @GT-center coverage is highest for the smallest lesions ($0.994$ below $6$\,mm) and falls with size ($0.869$ above $30$\,mm): a box regressor on a $36\times36$ grid resolves a compact target more easily than a large irregular one. Chained, the ordering reverses in the smallest bin ($0.726$ below $6$\,mm against $0.862$ at $6$--$20$\,mm), because localization rather than regression is the limiting factor exactly where lesions are smallest.

\paragraph{Slice presence.} Alongside localization the slice-level head predicts, per slice, whether a lesion is present (Table~\ref{tab:pres}).

\begin{table}[!htbp]
\centering
\small
\begin{tabular}{lrrrrrr}
\hline
Backbone & \#Slices & Prev. & AUROC & $F_1$ & P & R\\
\hline
SALT ViT-L  & 121{,}755 & $8.5\%$ & \textbf{0.956} & 0.663 & 0.588 & 0.759\\
DINOv2$+$reg (natural)     & 121{,}755 & $8.5\%$ & 0.949 & 0.632 & 0.561 & 0.722\\
SALT ViT-B  & 121{,}755 & $8.5\%$ & 0.931 & 0.481 & 0.339 & 0.832\\
Curia                      & 121{,}755 & $8.5\%$ & 0.951 & 0.622 & 0.523 & 0.768\\
MedDINOv3                  & 121{,}755 & $8.5\%$ & 0.942 & 0.630 & 0.639 & 0.621\\
DINO-LG ViT-B       & 121{,}755 & $8.5\%$ & 0.934 & 0.600 & 0.589 & 0.612\\
Curia-2                    & 121{,}755 & $8.5\%$ & 0.914 & 0.543 & 0.513 & 0.577\\
\hline
\end{tabular}
\caption{Slice-presence classifier (does a slice contain a lesion): slice count, positive prevalence, AUROC, and
$F_1$/precision/recall at threshold $0.5$. Prevalence is $8.5\%$; AUROC is the threshold-free ranking measure and the
operative quantity for the gate. Protocol and result below.}
\label{tab:pres}
\end{table}

\textbf{How it is measured.} Over all test slices we score the slice head's ``lesion present'' logit against the binary slice label. AUROC is threshold-free and measures ranking quality; $F_1$/precision/recall are read at a fixed $0.5$ probability threshold, the training-time default. In the gated detection pipeline of Table~\ref{tab:det} the gate is instead swept together with the detection threshold, so AUROC is the operative quantity there; the fixed-$0.5$ figures depend on each backbone's score calibration and do not affect the gated detection numbers.

\textbf{Result.} The classifier is a strong lesion-slice gate for every backbone (AUROC $0.914$--$0.956$), our SALT ViT-L leading at $0.956$. At the fixed $0.5$ threshold the calibration varies (our ViT-B, for instance, favours recall, $0.832$), which is why AUROC rather than $F_1$ is the reported ranking. The band across backbones is narrow, so the gate is not the component that distinguishes the encoders.

\subsection{Detection-free longitudinal re-identification}
A lesion-aware encoder should let us re-identify a known lesion in a later scan. We test this directly on the HCUCH RECIST longitudinal data, over $28$ lung and lymph-node target lesions, in a probe that involves no detector at all: the query is seeded from the ground-truth baseline center and matched against the follow-up volume by cosine similarity alone, so the result reflects the frozen representation rather than any trained head (Table~\ref{tab:find}).

\begin{table}[!htbp]
\centering
\small
\begin{tabular}{lrrrrl}
\hline
Backbone & find@$1$ & find@$3$ & find@$5$ & med.\ (mm) & find@$1$ $95\%$ CI\\
\hline
SALT ViT-B  & \textbf{0.679} & \textbf{0.857} & \textbf{0.893} & 10.4 & $[0.49, 0.82]$\\
Curia                      & 0.643 & 0.750 & 0.750 & \textbf{8.8} & $[0.46, 0.79]$\\
MedDINOv3                  & 0.643 & 0.750 & 0.750 & 12.2 & $[0.46, 0.79]$\\
DINO-LG ViT-B       & 0.571 & 0.679 & 0.714 & 13.2 & $[0.39, 0.73]$\\
Curia-2                    & 0.571 & 0.607 & 0.643 & 13.7 & $[0.39, 0.73]$\\
SALT ViT-L  & 0.536 & 0.750 & 0.786 & 14.6 & $[0.36, 0.70]$\\
DINOv2$+$reg (natural)     & 0.536 & 0.679 & 0.679 & 15.9 & $[0.36, 0.70]$\\
\hline
\end{tabular}
\caption{Detection-free lesion finding on HCUCH RECIST ($28$ lung/lymph-node target lesions). The baseline
lesion's single center-patch embedding is searched against the whole follow-up volume (every slice, every patch)
by cosine similarity; no detector head, registration, or follow-up annotation. find@$k$: correct lesion among
the top-$k$ similarity peaks; med.: $3$D median center error (mm) of the \#$1$ peak. Wilson $95\%$ CI on
find@$1$; all intervals overlap. Chance for a single whole-volume patch is $\ll0.05$ (thousands of candidate
patches). Protocol and result below.}
\label{tab:find}
\end{table}

\begin{figure}[!htbp]
\centering
\includegraphics[width=0.32\linewidth]{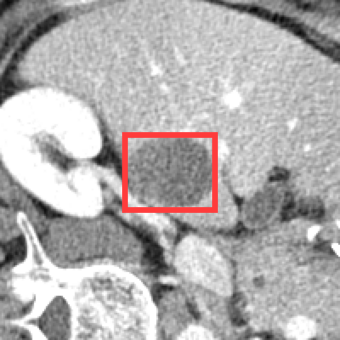}\hfill
\includegraphics[width=0.32\linewidth]{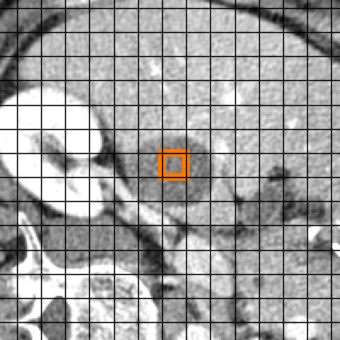}\hfill
\includegraphics[width=0.32\linewidth]{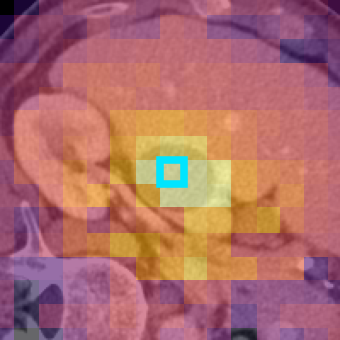}
\caption{Detection-free re-identification example (frozen SALT ViT-L; patient~$8$, abdomen). Left
to right: baseline CT with lesion box, query center patch on the $36\times36$ grid, and the follow-up
cosine-similarity map (described in the text).}
\label{fig:find}
\end{figure}

\textbf{How it is measured.} On the HCUCH RECIST cohort a target lesion is a physical lesion measured at both a baseline and a follow-up study (RECIST target lesions; $28$ lung/lymph-node lesions across the paired studies). For each lesion we take the single patch embedding at its baseline center, the query fingerprint (Figure~\ref{fig:find}, middle), and search the \emph{entire} follow-up volume, scoring every patch on every slice by cosine similarity to the query and extracting the top-$5$ similarity peaks by $3$D non-maximum suppression. A find is counted when a peak lands inside the true follow-up lesion. We report find@$k$ (the correct lesion is among the top-$k$ peaks) and the $3$D median center error (mm) of the \#$1$ peak. This is deliberately harder than closed-set matching: the query must be localized among thousands of candidate patches spanning the whole volume, and every lesion still draws $\sim\!3$ off-lesion peaks into its top-$5$, so the score reflects genuine retrieval against many distractors.

\textbf{Result.} The ordering here again favours target-level conditioning, though the sample is too small to establish it. Our SALT ViT-B finds the lesion at find@$1$ $0.679$ and find@$5$ $0.893$, a $+0.11$/$+0.18$ jump over the architecturally identical DINO-LG ViT-B, which shares its label-guided cropping but conditions no targets ($0.571$/$0.714$), and is the strongest backbone here, ahead of the private CT foundation model Curia and the public MedDINOv3 (both find@$1$ $0.643$) and well clear of the natural-image and Curia-2 references. Curia localizes most tightly when it does hit ($8.8$\,mm median error). These gaps are a few lesions wide: the Wilson intervals overlap for every pair, including our SALT ViT-B ($[0.49,0.82]$) against its DINO-LG counterpart ($[0.39,0.73]$), and the supervised ViT-L detector's edge does not carry over to this single-patch zero-shot retrieval. We therefore read the experiment as a proof of concept that the frozen conditional features alone re-identify a lesion across time, consistent with the patch geometry of Table~\ref{tab:sep}, and not as a second detector ranking. Figure~\ref{fig:find} illustrates the mechanism on one lesion (patient~$8$, abdomen); the three panels share a field of view. The left panel is the baseline CT with the target-lesion box; the middle overlays the $36\times36$ ViT patch grid on the same CT and marks the lesion's center patch, the query fingerprint, in orange; the right panel colors every patch of the follow-up volume by its cosine similarity to that query patch (inferno, brighter is more similar, no thresholding), with the single most-similar patch outlined in cyan. The similarity is highest exactly on the same physical lesion (peak cosine $0.94$), re-identifying it with no detector, mask, or fine-tuning.

\section{Discussion \& Conclusion}
\label{sec:discussion}

The result that organizes this paper is the gap between two encoders that differ in one thing. Both are ViT-B/14, trained on the same eight-cohort store with the same schedule, seed and lesion-guided cropping; one conditions its distillation targets on the lesion region and the other does not. Their separation margins are $0.449$ and $0.105$, and their pooled CPMs are $0.508$ and $0.394$. The mechanism claim rests on that pair; the external backbones place it in context.

Moving the guidance from the view to the target addresses a scale mismatch. Label-guided cropping makes lesions appear more often among the views, but a global view is $224\times224$ while a lesion is two or three patches across, so within any conditioned crop the objective still spends nearly all of its capacity on the surrounding parenchyma: the lesion is present but not privileged. Sharpening the teacher's target changes instead what the student must reproduce, and does so at exactly the implicated positions. Two consequences follow, and both are observed. The effect appears in the representation before any detector is involved, as a fourfold change in separation margin. And it is largest where the mismatch is most severe: the SALT ViT-B gains $+0.136$ CPM on sub-$6$\,mm lesions and is indistinguishable from its counterpart between $6$ and $20$\,mm (Table~\ref{tab:vitb_size}).

The benefit is not uniform across cohorts. Conditioning lifts CPM by $+0.144$ on UniToChest and $+0.210$ on NIH-Lymph and is slightly negative on NLST-seg and HCUCH, the two cohorts on which every backbone already performs comparably and where little headroom remains (Sec.~\ref{sec:cohortres}). Four cohorts, one pretraining seed and no interval on the paired difference cannot separate a headroom ceiling from genuine anatomy-dependence; a broader cohort panel would.

The external comparison separates two contributions that the matched pair does not. The natural-image DINOv2 ViT-L is the second-strongest detector despite an unremarkable patch geometry, localizing the most lesions of any backbone ($90\%$) while delineating them worst ($0.647$ chained coverage). Capacity and general visual pretraining therefore supply localization ability, while the conditioning supplies selectivity. Because those backbones differ from ours in anatomy coverage, patch size and recipe, they position the work rather than establish the mechanism.

Three limits bound what these numbers mean. Detection is scored catch-once, which mirrors the clinical read but does not score volumetric delineation; the detector is trained and evaluated on lung and lymph-node lesions only; and the longitudinal probe rests on $28$ paired lesions, too few to rank encoders. Within the mechanism, the sharpened temperature and the up-weighted patch loss remain unseparated, and a sweep over both would resolve their individual contributions.

Conditioning the teacher's target on a compact region derived from weak, pretraining-only labels turns a linearly decodable lesion direction into a geometrically concentrated one, and that change survives into detection through a frozen encoder and a head trained the same way for every backbone. Because the conditioning is expressed through a spatial indicator rather than through label semantics, the formulation admits any weak spatial annotation; validating it beyond the lesion instantiation is the natural next step.

\clearpage
\bibliographystyle{unsrt}
\bibliography{references}

\end{document}